%% file: varying.tex
\documentclass[]{article}
\usepackage[T1]{fontenc}
\usepackage{colacl}
\usepackage{rotating}    % to rotate them - only visible in .ps file
\usepackage{amssymb}     % AMS symbols
\usepackage{epsfig}      % postscript figures
\usepackage{verbatim}    % to comment regions
\usepackage{qtree}       % for parse trees

\input{macros}

\title{Bagging and Boosting a Treebank Parser}
\author{
  {\bf John C. Henderson} \\
  The MITRE Corporation \\
  202 Burlington Road \\
  Bedford, MA 01730 \\
  jhndrsn@mitre.org
  \And 
  {\bf Eric Brill} \\
  Microsoft Research \\
  1 Microsoft Way \\
  Redmond, WA 98052 \\
  brill@microsoft.com 
}

\begin{document}
\maketitle

\begin{abstract}
 \begin{picture}(0,0)
 \put(0,215){Appears in}
 \put(5,200){Proceedings of the 1st Meeting of the North
   American Chapter of the Association for Computational Linguistics}
 \put(10,185){(NAACL-2000), pages 34--41.  Seattle, Washington, USA,
   April--May, 2000.}
 \end{picture}
  Bagging and boosting, two effective machine learning techniques, are
  applied to natural language parsing.  Experiments using these
  techniques with a trainable statistical parser are described.  The
  best resulting system provides roughly as large of a gain in
  F-measure as doubling the corpus size.  Error analysis of the result
  of the boosting technique reveals some inconsistent annotations in
  the Penn Treebank, suggesting a semi-automatic method for finding
  inconsistent treebank annotations.
\end{abstract}

\section{Introduction}

\newcite{henderson:parsercombo} showed that independent human research
efforts produce parsers that can be combined for an overall boost in
accuracy.  Finding an ensemble of parsers designed to complement each
other is clearly desirable.  The parsers would need to be the result
of a unified research effort, though, in which the errors made by one
parser are targeted with priority by the developer of another parser.

A set of five parsers which each achieve only 40\% exact sentence
accuracy would be extremely valuable if they made errors in such a way
that at least two of the five were correct on any given sentence (and
the others abstained or were wrong in different ways).  100\% sentence
accuracy could be achieved by selecting the hypothesis that was
proposed by the two parsers that agreed completely.

In this paper, the task of automatically creating complementary
parsers is separated from the task of creating a single parser.  This
facilitates study of the ensemble creation techniques in isolation.
The result is a method for increasing parsing performance by creating
an ensemble of parsers, each produced from data using the same parser
induction algorithm.

\section{Bagging and Parsing}

\subsection{Background}

The work of \newcite{efron93:bootstrap} enabled Breiman's refinement
and application of their techniques for machine learning
\cite{breiman96:bagging}.  His technique is called \emph{bagging},
short for ``bootstrap aggregating''.  In brief, bootstrap techniques
and bagging in particular reduce the systematic biases many estimation
techniques introduce by aggregating estimates made from randomly
drawn representative resamplings of those datasets.

Bagging attempts to find a set of classifiers which are consistent
with the training data, different from each other, and distributed
such that the aggregate sample distribution approaches the
distribution of samples in the training set.

%The bagging algorithm is shown in Algorithm \ref{algorithm:rawbagging}.

\begin{algorithm}
{Bagging Predictors \\ (Breiman, 1996)}
{Given: training set $\mathcal L = \{(y_i,x_i), i \in \{1\ldots
  m\}\}$ drawn from the set $\Lambda$ of possible training sets where 
  $y_i$ is the label for example $x_i$, classification induction
  algorithm $\Psi:\Lambda \rightarrow \Phi$ with classification
  algorithm $\phi \in \Phi$ and $\phi:X\rightarrow Y$.}
\label{algorithm:rawbagging}

\item Create $k$ bootstrap replicates of $\mathcal L$ by sampling $m$
  items from $\mathcal L$ \emph{with replacement}.  Call them
  $L_1 \ldots L_k$.
  
\item For each $j\in \{1\ldots k\}$, Let $\phi_j = \Psi(\mathcal L_j)$
  be the classifier induced using $L_j$ as the training set.
  
\item If $Y$ is a discrete set, then for each $x_i$ observed in the
  test set, $y_i = \mathrm{mode} \langle \phi_j(x_i)\ldots
  \phi_j(x_i)\rangle$.  $y_i$ is the value predicted
  by the most predictors, the majority vote.%\footnotemark
\end{algorithm}
% \footnotetext{
%   When the range, $Y$, is an interval in $\Re$, the regression form of
%   bagging uses the arithmetic mean, $y_i = \frac{1}{k}\sum\limits_j
%   \phi_j(x_i)$.  }

\begin{comment}
  There are two interesting qualitative properties of bagging.  First,
  bagging relies on the chosen classifier induction algorithm's lack
  of \emph{stability}.  This means the chosen algorithm should be
  easily perturbed.  A small change in the training set should produce
  a significant change in the resulting classifier.  Neural networks
  and decision trees are examples of unstable classifier systems,
  whereas k-nearest neighbor is a stable classifier.  Secondly,
  bagging is theoretically resistant to noise in the data and bias in
  the learning algorithm.  Unfortunately it is resistant to bias in
  the learning algorithm even when that bias is favorable. Classifier
  induction algorithms that perform well in isolation can perform
  poorly in ensemble for this reason.  Empirical results have verified
  both of these claims
  \cite{quinlan96:bagging_boosting,maclin97:baggingvboosting,bauer99:bagboost}.
\end{comment}

\subsection{Bagging for Parsing}

An algorithm that applies the technique of bagging to parsing is given
in Algorithm \ref{algorithm:parsebagging}.  Previous work on combining
independent parsers is leveraged to produce the combined parser.  The
rest of the algorithm is a straightforward transformation of bagging
for classifiers.  Exploratory work in this vein was described by
\newcite{hajic:ws98}.

\begin{algorithm}
{Bagging A Parser}
{Given: A corpus (again as a function) $\mathcal C:S\times T \rightarrow N$,
  $S$ is the set of possible sentences, and $T$ is the set of trees,
  with size $m=|\mathcal C|=\sum_{s,t}\mathcal C(s,t)$ and parser induction algorithm
  $g$.} 
\label{algorithm:parsebagging}
\item Draw $k$ bootstrap replicates $C_1\ldots C_k$ of $\mathcal C$
  each containing $m$ samples of $(s,t)$ pairs randomly picked from
  the domain of $\mathcal C$ according to the distribution $D(s,t) =
  \mathcal C(s,t)/|\mathcal C|$.  Each bootstrap replicate is a bag of samples,
  where each sample in a bag is drawn randomly with replacement from
  the bag corresponding to $\mathcal C$.
\item Create parser $f_i = g(C_i)$ for each $i$.
% $F_{ensemble} =  \bigcup_i \{f_i\}$.
\item Given a novel sentence $s_{test}\in \mathcal C_{test}$, combine
  the collection of hypotheses $ t_i =f_i(s_{test})$ using the
  unweighted constituent voting scheme of
  \newcite{henderson:parsercombo}.
\end{algorithm}

\subsection{Experiment}

The training set for these experiments was sections 01-21 of the Penn
Treebank \cite{penntreebank}.  The test set was section 23.  The
parser induction algorithm used in all of the experiments in this
paper was a distribution of Collins's model 2
parser \cite{collins:parsing97}.  All comparisons made below refer to
results we obtained using Collins's parser.

\begin{table*}[htbp]
  \centering
  \input{bag.sent.uni.tab.tex}
  \caption{Bagging the Treebank} 
  \label{table:bag.sent.uni}
\end{table*}

The results for bagging are shown in Figure \ref{fig:bag.sent.uni} and
Table \ref{table:bag.sent.uni}.  The row of figures are (from
left-to-right) training set F-measure\footnote{This is the balanced
  version of F-measure, where precision and recall are weighted
  equally.}, test set F-measure, percent perfectly parsed sentences in
training set, and percent perfectly parsed sentences in test set.  An
ensemble of bags was produced one bag at a time.  In the table, the
{\tt Initial} row shows the performance achieved when the ensemble
contained only one bag, {\tt Final(X)} shows the performance when the
ensemble contained $X$ bags, {\tt BestF} gives the performance of the
ensemble size that gave the best F-measure score.  {\tt TrainBestF}
and {\tt TestBestF} give the test set performance for the ensemble
size that performed the best on the training and test sets,
respectively.

On the training set all of the accuracy measures are improved over the
original parser, and on the test set there is clear improvement in
precision and recall.  The improvement on exact sentence accuracy for
the test set is significant, but only marginally so.

The overall gain achieved on the test set by bagging was 0.8 units of
F-measure, but because the entire corpus is not used in each bag the
initial performance is approximately 0.2 units below the best
previously reported result.  The net gain using this technique is 0.6
units of F-measure.

\section{Boosting}

\subsection{Background}

The AdaBoost algorithm was presented by Freund and Schapire in 1996
\cite{freund96:adaboost_experiments,freund97:adaboost} and has become
a widely-known successful method in machine learning.
% There is
% prior work on boosting \cite{schapire90,freund95:boostmaj}, but
% the discussion in this paper follows the AdaBoost work. 
The AdaBoost algorithm imposes one constraint on its underlying
learner: it may abstain from making predictions about labels of some
samples, but it must consistently be able to get more than 50\%
accuracy on the samples for which it commits to a decision.  That
accuracy is measured according to the distribution describing the
importance of samples that it is given.  The learner must be able to
get more correct samples than incorrect samples \emph{by mass of
  importance} on those that it labels.  This statement of the
restriction comes from Schapire and Singer's study
\shortcite{ss:boost98:colt}.  It is called the \emph{weak learning
  criterion}.

Schapire and Singer \shortcite{ss:boost98:colt} extended AdaBoost by
describing how to choose the hypothesis mixing coefficients in certain
circumstances and how to incorporate a general notion of confidence
scores.  They also provided a better characterization of its
theoretical performance.  The version of AdaBoost used in their work
is shown in Algorithm \ref{algorithm:adaboost}, as it is the version
that most amenable to parsing.

\begin{algorithm}
  {AdaBoost \\ (Freund and Schapire, 1997)}
  {Given: Training set $\mathcal L$ as in bagging, except $y_i \in \{-1,1\}$ is the label for example $x_i$.
    Initial uniform distribution $D_1(i) = 1/m$.  Number of
    iterations, $T$. Counter $t=1$.  $\Psi$, $\Phi$, and $\phi$ are as
    in Bagging.}
\label{algorithm:adaboost}

\item Create $L_t$ by randomly choosing with replacement $m$ samples
 from $\mathcal L$ using distribution $D_t$.\label{algorithm:adaboost:loopstart}

\item Classifier induction: $\phi_t \leftarrow \Psi(L_t)$

\item Choose $\alpha_t \in \Re$.

\item Adjust and normalize the distribution.  $Z_t$ is a normalization coefficient. 
\begin{displaymath}
  D_{t+1}(i) = \frac{1}{Z_t}{D_t(i)\exp(-\alpha_t y_i \phi_t(x_i))}
\end{displaymath}

\item Increment $t$.  Quit if $t > T$.

\item Repeat from step \ref{algorithm:adaboost:loopstart}.
\item The final hypothesis is 
\begin{displaymath}
  \phi_{boost}(x) = \mathrm{sign} \sum_t  \alpha_t\phi_t(x)
\end{displaymath}
\end{algorithm}

The value of $\alpha_t$ should generally be chosen to minimize
\begin{displaymath}
  \sum_i D_t(i)\exp(-\alpha_t y_i \phi_t(x_i))
\end{displaymath}
in order to minimize the expected per-sample training error of the
ensemble, which Schapire and Singer show can be concisely expressed by
$\prod\limits_t Z_t$.  They also give several examples for
how to pick an appropriate $\alpha$, and selection generally depends
on the possible outputs of the underlying learner.

Boosting has been used in a few NLP systems.
\newcite{boostingparsing98} used boosting to produce more
accurate classifiers which were embedded as control mechanisms of a
parser for Japanese.  The creators of AdaBoost used it to perform text
classification \cite{boosttext2000}.  \newcite{abney99:boosttagpp}
performed part-of-speech tagging and prepositional phrase attachment
using AdaBoost as a core component.  They found they could achieve
accuracies on both tasks that were competitive with the state of the
art. As a side effect, they found that inspecting the samples that
were consistently given the most weight during boosting revealed some
faulty annotations in the corpus.  In all of these systems, AdaBoost
has been used as a traditional classification system.

\subsection{Boosting for Parsing}

Our goal is to recast boosting for parsing while considering a parsing
system as the embedded learner.  The formulation is given in Algorithm
\ref{algorithm:parseboosting}.  The intuition behind the additive form
is that the weight placed on a sentence should be the sum of the
weight we would like to place on its constituents.  The weight on
constituents that are predicted incorrectly are adjusted by a factor
of 1 in contrast to a factor of $\alpha$ for those that are predicted
incorrectly.

\begin{algorithm}
{Boosting A Parser}
{Given corpus $\mathcal C$ with size $m=|\mathcal
  C|=\sum_{s,\tau}\mathcal C(s,t)$ and
  parser induction algorithm $g$.    Initial uniform distribution $D_1(i)
    = 1/m$.  Number of iterations, $T$. Counter $t=1$.}
\label{algorithm:parseboosting}
\item Create $C_t$ by randomly choosing with replacement $m$
  samples from $\mathcal C$ using distribution
  $D_t$.\label{algorithm:boostparse:loopstart}
\item Create parser $f_t \leftarrow g(C_t)$.
\item Choose $\alpha_t \in \Re$ (described below).\label{algorithm:boostparse:choosealpha}
  
\item Adjust and normalize the distribution.  $Z_t$ is a normalization
  coefficient.  For all $i$, let parse tree $\tau^\prime_i \leftarrow
  f_t(s_i)$.  Let $\delta(\tau,c)$ be a function indicating that $c$
  is in parse tree $\tau$, and $|\tau|$ is the number of constituents
  in tree $\tau$.  $T(s)$ is the set of constituents that are
  found in the reference or hypothesized annotation for
  $s$.\label{algorithm:parseboosting:tricky}
{\small
\begin{eqnarray*}
\lefteqn{  D_{t+1}(i) =} \nonumber\\
&&\frac{1}{Z_t}{D_t(i)  \sum\limits_{c\in T(s_i)}
\left(\alpha +
      (1-\alpha)|\delta(\tau^\prime_i,c)- \delta(\tau_i,c)|\right)}
\end{eqnarray*}
}
\item Increment $t$.  Quit if $t > T$.

\item Repeat from step \ref{algorithm:boostparse:loopstart}.
  
\item The final hypothesis is computed by combining the individual
  constituents.  Each parser $\phi_t$ in the ensemble gets a vote with
  weight $\alpha_t$ for the constituents they predict.  Precisely
  those constituents with weight strictly larger than
  $\frac{1}{2}\sum_t\alpha_t$ are put into the final hypothesis.

\end{algorithm}

A potential constituent can be considered correct if it is predicted
in the hypothesis and it exists in the reference, or it is not
predicted and it is not in the reference.  Potential constituents that
do not appear in the hypothesis or the reference should not make a big
contribution to the accuracy computation.  There are many such
potential constituents, and if we were maximizing a function that
treated getting them incorrect the same as getting a constituent that
appears in the reference correct, we would most likely decide not to
predict any constituents.

Our model of constituent accuracy is thus simple.  Each prediction
correctly made over $T(s)$ will be given equal weight.  That is,
correctly hypothesizing a constituent in the reference will give us
one point, but a precision or recall error will cause us to miss one
point.  Constituent accuracy is then $a/(a+b+c)$, where $a$ is the
number of constituents correctly hypothesized, $b$ is the number of
precision errors and $c$ is the number of recall errors.

In Equation \ref{eqn:alphabca}, a computation of $\alpha_{ca}$
as described is shown.

{\small
\begin{eqnarray}
\alpha_{ca} =
\frac{
  \sum\limits_i \frac{D(i)}{|T(s_i)|}
  \sum\limits_{c\in T(s_i)}
    \delta(\tau_i,c)
    +\delta(\tau^\prime_i,c)
    -2\delta(\tau_i,c)\delta(\tau^\prime_i,c)
}
{
  \sum\limits_i \frac{D(i)}{|T(s_i)|}
  \sum\limits_{c\in T(s_i)}\delta(\tau_i,c)\delta(\tau^\prime_i,c)
}
\label{eqn:alphabca}
\end{eqnarray}
}

Boosting algorithms were developed that attempted to maximize
F-measure, precision, and recall by varying the computation of
$\alpha$, giving results too numerous to include here.  The algorithm
given here performed the best of the lot, but was only marginally
better for some metrics.

\subsection{Experiment}

\begin{table*}[htbp]
  \centering
  \input{boost.sent.tab.tex}
  \caption{Boosting the Treebank} 
  % includes rebagging
  \label{table:boost.sent}
\end{table*}

The experimental results for boosting are shown in Figure
\ref{fig:boost.sent} and Table \ref{table:boost.sent}.
%The first thing to notice is that the notch in all the
%graphs at iteration 13 comes from the boosting algorithm backing off
%to bagging on that iteration.
There is a large plateau in performance from iterations 5 through 12.
Because of their low accuracy and high degree of specialization, the
parsers produced in these iterations had little weight during voting
and had little effect on the cumulative decision making.

As in the bagging experiment, it appears that there would be more
precision and recall gain to be had by creating a larger ensemble.  In
both the bagging and boosting experiments time and resource
constraints dictated our ensemble size.

In the table we see that the boosting algorithm equaled bagging's test
set gains in precision and recall.  The {\tt Initial} performance for
boosting was lower, though.  We cannot explain this, and expect it is
due to unfortunate resampling of the data during the first iteration
of boosting.  Exact sentence accuracy, though, was not significantly
improved on the test set.

Overall, we prefer bagging to boosting for this problem when raw
performance is the goal.  There are side effects of boosting that
are useful in other respects, though, which we explore in Section
\ref{section:qualitycontrol}.

\subsubsection{Weak Learning Criterion Violations}
\label{section:weaklearn}

It was hypothesized in the course of investigating the failures of the
boosting algorithm that the parser induction system did not satisfy
the weak learning criterion.  It was noted that the distribution of
boosting weights were more skewed in later iterations.  Inspection of
the sentences that were getting much mass placed upon them revealed
that their weight was being boosted \emph{in every iteration}.  The
hypothesis was that the parser was simply unable to learn them.

39832 parsers were built to test this, one for each sentence
in the training set.  Each of these parsers was trained on only a
single sentence\footnote{The sentence was replicated 10 times to avoid
  thresholding effects in the learner.} and evaluated on the same
sentence.  It was discovered that a full 4764 (11.2\%) of these
sentences could not be parsed completely correctly by the parsing
system.

\subsubsection{Corpus Trimming}
\label{section:trimming}

\begin{table*}[htbp]
\centering
  \input{boost.stable.tab.tex}
  \caption{Boosting the Stable Corpus} 
  \label{table:boost.stable}
\end{table*}

In order to evaluate how well boosting worked with a learner that
better satisfied the weak learning criterion, the boosting experiment
was run again on the Treebank minus the troublesome sentences
described above.  The results are in Table \ref{table:boost.stable}.
This dataset produces a larger gain in comparison to the results using
the entire Treebank.  The initial accuracy, however, is lower.  We
hypothesize that the boosting algorithm did perform better here, but
the parser induction system was learning useful information in those
sentences that it could not memorize (e.g. lexical information) that
was successfully applied to the test set.

In this manner we managed to clean our dataset to the point that the
parser could learn each sentence in isolation.  The corpus-makers
cannot necessarily be blamed for the sentences that could not be
memorized.  All that can be said about those sentences is that for
better or worse, the parser's model would not accommodate them.

\section{Corpus Analysis}

\subsection{Noisy Corpus: Empirical Investigation}

To acquire experimental evidence of noisy data, distributions that
were used during boosting the stable corpus were inspected. The
distribution was expected to be skewed if there was noise in the data,
or be uniform with slight fluctuations if it fit the data well.

\begin{figure}
  \centering
  \epsfig{file=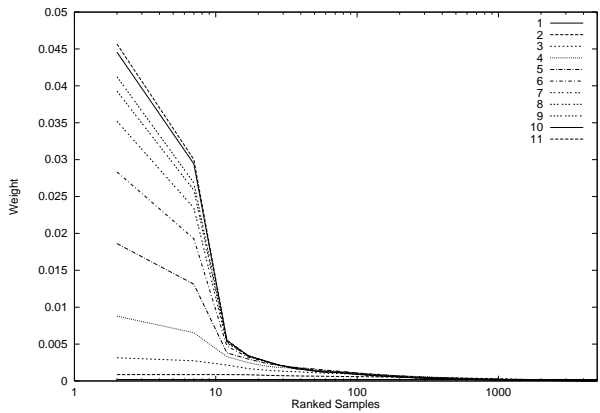, width=\columnwidth}
  \caption{Weight Change During Boosting}
  \label{figure:freqrank} 
\end{figure}

We see how the boosting weight distribution changes in Figure
\ref{figure:freqrank}.  The individual curves are indexed by boosting
iteration in the key of the figure.  This training run used a corpus
of 5000 sentences. The sentences are ranked by the weight they are
given in the distribution, and sorted in decreasing order by weight
along the x-axis.  The distribution was smoothed by putting samples
into equal weight bins, and reporting the average mass of samples in
the bin as the y-coordinate.  Each curve on this graph corresponds to
a boosting iteration.  We used 1000 bins for this graph, and a log
scale on the x-axis.  Since there were 5000 samples, all samples
initially had a y-value of 0.0002.

Notice first that the left endpoints of the lines move from bottom to
top in order of boosting iteration.  The distribution becomes
monotonically more skewed as boosting progresses.  Secondly we see by
the last iteration that most of the weight is concentrated on less
than 100 samples. This graph shows behavior consistent with noise in
the corpus on which the boosting algorithm is focusing.

\subsection{Treebank Inconsistencies}
\label{section:qualitycontrol}

There are sentences in the corpus that can be learned by the parser
induction algorithm in isolation but not in concert because they
contain conflicting information.  Finding these sentences leads to a
better understanding of the quality of our corpus, and gives an idea
for where improvements in annotation quality can be made.
\newcite{abney99:boosttagpp} showed a similar corpus analysis
technique for part of speech tagging and prepositional phrase tagging,
but for parsing we must remove errors introduced by the parser as we
did in Section \ref{section:trimming} before questioning the corpus
quality.  A particular class of errors, inconsistencies, can then be
investigated.  Inconsistent annotations are those that appear
plausible in isolation, but which conflict with annotation decisions
made elsewhere in the corpus.

In Figure \ref{fig:treebankerrors} we show a set of trees selected
from within the top 100 most heavily weighted trees at the end of 15
iterations of boosting the stable corpus.%\footnote{More examples are
%  available from the authors upon request.}
Collins's parser induction system is able to
learn to produce any one of these structures in isolation, but the
presence of conflicting information \emph{in different sentences}
prevents it from achieving 100\% accuracy on the set.

% Within the top 100 sentences, there were also some trees that did not
% appear to have obvious problems.  There are a few potential causes for
% this: the boosting algorithm was prematurely stopped, there could be
% an overabundance (majority) of incorrectly annotated trees further
% down in the ranking, or the trees may expose inadequacies of the
% parsing model.

\section{Training Corpus Size Effects}

We suspect our best parser diversification techniques gives
performance gain approximately equal to doubling the size of the
training set.  While this cannot be directly tested without hiring
more annotators, an expected performance bound for a larger training
set can be produced by extrapolating from how well the parser performs
using smaller training sets.  There are two characteristics of
training curves for large corpora that can provide such a bound:
training curves generally increase monotonically in the absence of
over-training, and their first derivatives generally decrease
monotonically.

{\small
\begin{table}[htbp]
  \begin{center}
    \begin{tabular}{|c|l|rr|r|r|}
      \hline 
      Set
      &\multicolumn{1}{|c|}{Sentences}
      &\multicolumn{1}{c}{P} 
      & \multicolumn{1}{c|}{R} 
      & \multicolumn{1}{c|}{F}  
      & \multicolumn{1}{c|}{Exact}
      \\
      \hline 
% results when run on entire training corpus
      &50    & 67.57 & 32.15 & 43.57 &  5.4 \\
      &100   & 69.03 & 56.23 & 61.98 &  8.5 \\
      &500   & 78.12 & 75.46 & 76.77 & 18.2 \\
      &1000  & 81.36 & 80.70 & 81.03 & 22.9 \\
      &5000  & 87.28 & 87.09 & 87.19 & 34.1 \\
      &10000 & 89.74 & 89.56 & 89.65 & 41.0 \\
      &20000 & 92.42 & 92.40 & 92.41 & 50.3 \\
      \begin{rotate}{90}Training\end{rotate}
      &39832 & 96.25 & 96.31 & 96.28 & 64.7 \\
      \hline
% results when run on test corpus
      &50    & 68.13 & 32.24 & 43.76  &  4.7 \\
      &100   & 69.90 & 54.19 & 61.05  &  7.8 \\
      &500   & 78.72 & 75.33 & 76.99  & 19.1 \\
      &1000  & 81.61 & 80.68 & 81.14  & 22.2 \\
      &5000  & 86.03 & 85.43 & 85.73  & 28.6 \\
      &10000 & 87.29 & 86.81 & 87.05  & 30.8 \\
      &20000 & 87.99 & 87.87 & 87.93  & 32.7 \\
      \begin{rotate}{90}Testing\end{rotate}
      &39832 & 88.73 &   88.54 &   88.63  & 34.9 \\
      \hline 
    \end{tabular}
  \caption{Effects of Varying Training Corpus Size} 
  \label{table:sizes.train}
  \end{center}
\end{table}
}

The training curves we present in Figure \ref{fig:sizes} and
Table \ref{table:sizes.train} suggest that roughly doubling
the corpus size in the range of interest (between 10000 and 40000
sentences) gives a test set F-measure gain of approximately 0.70.

Bagging achieved significant gains of approximately 0.60 over the best
reported previous F-measure without adding any new data.  In this
respect, these techniques show promise for making performance gains on
large corpora without adding more data or new parsers.

% It should be noted that the parser induction system performs
% surprisingly well when training with only a small amount of data.
% This is support for why the feature-rich parsing model of Hermjakob
% and Mooney \cite{mooneyparse1997} can perform competitively with such
% a small quantity of data.
% It was not previously known that Collins's
% parser (which uses a knowledge-impoverished model by comparison) can
% achieve 90\% of its maximum performance using a training set of only
% 1000 sentences.

\section{Conclusion}

We have shown two methods, bagging and boosting, for automatically
creating ensembles of parsers that produce better parses than any
individual in the ensemble.  Neither of the algorithms exploit any
specialized knowledge of the underlying parser induction algorithm,
and the data used in creating the ensembles has been restricted to a
single common training set to avoid issues of training data quantity
affecting the outcome.

Our best bagging system performed consistently well on all metrics,
including exact sentence accuracy.  It resulted in a statistically
significant F-measure gain of 0.6 over the performance of the baseline
parser.  That baseline system is the best known Treebank parser.  This
gain compares favorably with a bound on potential gain from increasing
the corpus size.

Even though it is computationally expensive to create and evaluate a
small (15-30) ensemble of parsers, the cost is far outweighed by the
opportunity cost of hiring humans to annotate 40000 more sentences.
The economic basis for using ensemble methods will continue to improve
with the increasing value (performance per price) of modern hardware.

Our boosting system, although dominated by the bagging system, also
performed significantly better than the best previously known
individual parsing result.  We have shown how to exploit the
distribution created as a side-effect of the boosting algorithm to
uncover inconsistencies in the training corpus.  A semi-automated
technique for doing this as well as examples from the Treebank that
are inconsistently annotated were presented.  Perhaps the biggest
advantage of this technique is that it requires no a priori notion of
how the inconsistencies can be characterized.

\section{Acknowledgments}

We would like to thank Michael Collins for enabling all of this
research by providing us with his parser and helpful comments.

This work was funded by NSF grant IRI-9502312.  The views expressed in
this paper are those of the authors and do not necessarily reflect the
views of the MITRE Corporation.  This work was done while both authors
were at Johns Hopkins University.

\bibliographystyle{acl}
\bibliography{varying}
\onecolumn

\begin{sidewaysfigure}[htbp]
  \centering
  \input{bag.sent.uni.fig.tex}
  \caption{Bagging the Treebank} 
  \label{fig:bag.sent.uni}
\vspace{.25 in}
%\vspace{\vfill}
  \input{boost.sent.fig.tex}
  \caption{Boosting the Treebank} 
  % includes rebagging
  \label{fig:boost.sent}
%   \input{boost.stable.fig.tex}
%   \caption{Boosting the Stable Corpus} 
%   \label{fig:boost.stable}
  % weight/rank graphs
\vspace{.25 in}
%\vspace{\vfill}
  \input{sizes.vary.tex}
  \caption{Effects of Varying Training Corpus Size} 
  \label{fig:sizes}
\end{sidewaysfigure}

\input{treebankerrors}

\end{document}

%% file: macros.tex
\newcounter{parse}

{}
{}

\newcounter{algorithm}
\renewcommand{\thealgorithm}{\arabic{algorithm}}
\newenvironment{algorithm}[2]
{
\refstepcounter{algorithm}
\vspace{1em}\noindent
{\bf Algorithm: #1 \hfill (\thealgorithm)}

\noindent#2
\begin{enumerate}}
{\end{enumerate}}

\renewcommand{\Re}{\mathbb{R}} % preferred amstex

%% file: bag.sent.uni.tab.tex
%auto-ignore

{\small
\begin{tabular}{|c|l|rr|rr|rr|}
       \hline
      Set
      &\multicolumn{1}{c|}{Instance} 
      &\multicolumn{1}{c}{P} 
      &\multicolumn{1}{c|}{R} 
      &\multicolumn{1}{c}{F}  
      &\multicolumn{1}{c|}{Gain}  
      &\multicolumn{1}{c}{Exact}
      &\multicolumn{1}{c|}{Gain} \\
      \hline 
Training
&       Original Parser & 96.25 & 96.31 & 96.28 & NA    & 64.7 & NA
\\
&       Initial & 93.61 & 93.63 & 93.62 &  0.00 & 55.5  &  0.0
\\
&       BestF(15)       & 96.16 & 95.86 & 96.01 &  2.39 & 62.1  &  6.6
\\
&       Final(15)       & 96.16 & 95.86 & 96.01 &  2.39 & 62.1  &  6.6
\\
%\hline
%\end{tabular}
%\\
%\begin{tabular}{|l|rr|rr|rr|}
%       \hline
%      \multicolumn{1}{|c|}{Testing} 
%      &\multicolumn{1}{c}{P} 
%      &\multicolumn{1}{c|}{R} 
%      &\multicolumn{1}{c}{F}  
%      &\multicolumn{1}{c|}{Gain}  
%      &\multicolumn{1}{c}{Exact}
%      &\multicolumn{1}{c|}{Gain} \\
      \hline 
Test
& Original Parser &   88.73 &   88.54 &   88.63  & NA &34.9 &NA
\\
&   Initial & 88.43 & 88.34 & 88.38 &  0.00 & 33.3  &  0.0
\\
&       TrainBestF(15)  & 89.54 & 88.80 & 89.17 &  0.79 & 34.6  &  1.3
\\
&       TestBestF(13)   & 89.55 & 88.84 & 89.19 &  0.81 & 34.7  &  1.4
\\
&       Final(15)       & 89.54 & 88.80 & 89.17 &  0.79 & 34.6  &  1.3
\\
\hline
\end{tabular}
}
% \end{tabular}
%}
%\end{figure}

%%% Local Variables: 
%%% mode: latex
%%% TeX-master: "varying"
%%% End: 

%% file: boost.sent.tab.tex
%auto-ignore 

{\small
\begin{tabular}{|c|l|rr|rr|rr|}
       \hline
      Set
      &\multicolumn{1}{c|}{Instance} 
      &\multicolumn{1}{c}{P} 
      &\multicolumn{1}{c|}{R} 
      &\multicolumn{1}{c}{F}  
      &\multicolumn{1}{c|}{Gain}  
      &\multicolumn{1}{c}{Exact}
      &\multicolumn{1}{c|}{Gain} \\
      \hline 
Training
&       Original Parser & 96.25 & 96.31 & 96.28 & NA    & 64.7 & NA
\\
&       Initial & 93.54 & 93.61 & 93.58 &  0.00 & 54.8  &  0.0
\\
&       BestF(15)       & 96.21 & 95.79 & 96.00 &  2.42 & 57.3  &  2.5
\\
&       Final(15)       & 96.21 & 95.79 & 96.00 &  2.42 & 57.3  &  2.5
\\
%\hline
%\end{tabular}
%\\
%\begin{tabular}{|l|rr|rr|rr|}
%       \hline
%      \multicolumn{1}{|c|}{Testing} 
%      &\multicolumn{1}{c}{P} 
%      &\multicolumn{1}{c|}{R} 
%      &\multicolumn{1}{c}{F}  
%      &\multicolumn{1}{c|}{Gain}  
%      &\multicolumn{1}{c}{Exact}
%      &\multicolumn{1}{c|}{Gain} \\
      \hline 
Test
& Original Parser &   88.73 &   88.54 &   88.63  & NA &34.9 &NA
\\
&   Initial & 88.05 & 88.09 & 88.07 &  0.00 & 33.3  &  0.0
\\
&       TrainBestF(15)  & 89.37 & 88.32 & 88.84 &  0.77 & 33.0  & -0.3
\\
&       TestBestF(14)   & 89.39 & 88.41 & 88.90 &  0.83 & 33.4  &  0.1
\\
&       Final(15)       & 89.37 & 88.32 & 88.84 &  0.77 & 33.0  & -0.3
\\
\hline
\end{tabular}
}
% \end{tabular}
%\end{figure}

%%% Local Variables: 
%%% mode: latex
%%% TeX-master: "varying"
%%% End: 

%% file: boost.stable.tab.tex
%auto-ignore 

\small
\begin{tabular}{|c|l|rr|rr|rr|}
       \hline
      Set
      &\multicolumn{1}{c|}{Instance} 
      &\multicolumn{1}{c}{P} 
      &\multicolumn{1}{c|}{R} 
      &\multicolumn{1}{c}{F}  
      &\multicolumn{1}{c|}{Gain}  
      &\multicolumn{1}{c}{Exact}
      &\multicolumn{1}{c|}{Gain} \\
      \hline 
Training 
&       Original Parser & 96.25 & 96.31 & 96.28 & NA    & 64.7 & NA
\\
&       Initial         & 94.60 & 94.68 & 94.64 &  0.00 & 62.2  &  0.0
\\
&       BestF(8)        & 97.38 & 97.00 & 97.19 &  2.55 & 63.1  &  0.9
\\
&       Final(15)       & 97.00 & 96.17 & 96.58 &  1.94 & 55.0  & -7.2
\\
%\hline
%\end{tabular}
%\\
%\begin{tabular}{|l|rr|rr|rr|}
%       \hline
%      \multicolumn{1}{|c|}{Testing} 
%      &\multicolumn{1}{c}{P} 
%      &\multicolumn{1}{c|}{R} 
%      &\multicolumn{1}{c}{F}  
%      &\multicolumn{1}{c|}{Gain}  
%      &\multicolumn{1}{c}{Exact}
%      &\multicolumn{1}{c|}{Gain} \\
      \hline 
Test
& Original Parser &   88.73 &   88.54 &   88.63  & NA &34.9 &NA
\\
&   Initial & 87.43 & 87.21 & 87.32 &  0.00 & 32.6  &  0.0
\\
&       TrainBestF(8)   & 89.12 & 87.62 & 88.36 &  1.04 & 32.8  &  0.2
\\
&       TestBestF(6)    & 89.07 & 87.77 & 88.42 &  1.10 & 32.9  &  0.4
\\
&       Final(15)       & 89.18 & 87.19 & 88.18 &  0.86 & 31.7  & -0.8
\\
\hline
\end{tabular}
% \end{tabular}
%}
%\end{figure}

%%% Local Variables: 
%%% mode: latex
%%% TeX-master: "varying"
%%% End: 

%% file: bag.sent.uni.fig.tex
%auto-ignore

%\begin{figure}[htbp]
\mbox{
%\begin{tabular}{c}
\begin{tabular}{cccc}
    \epsfig{file=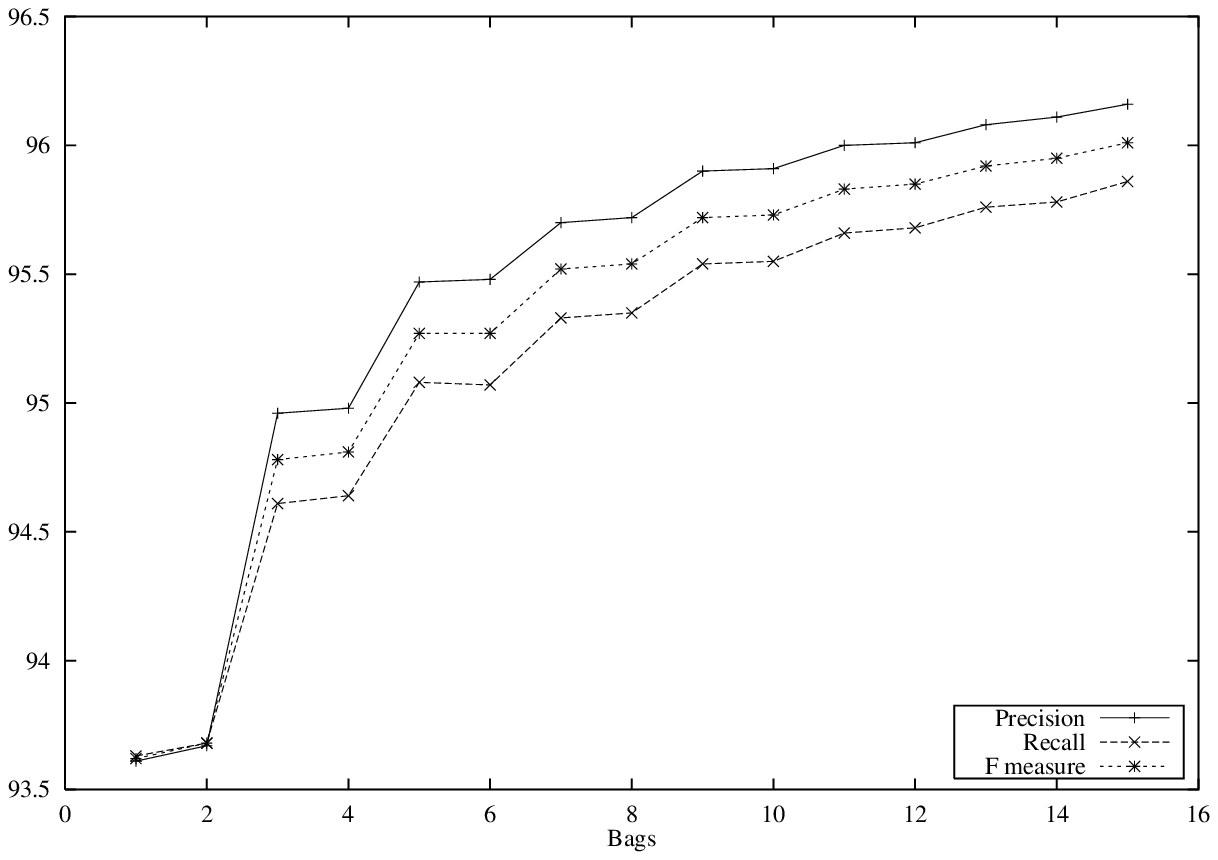, width= 0.25\textwidth}
&
    \epsfig{file=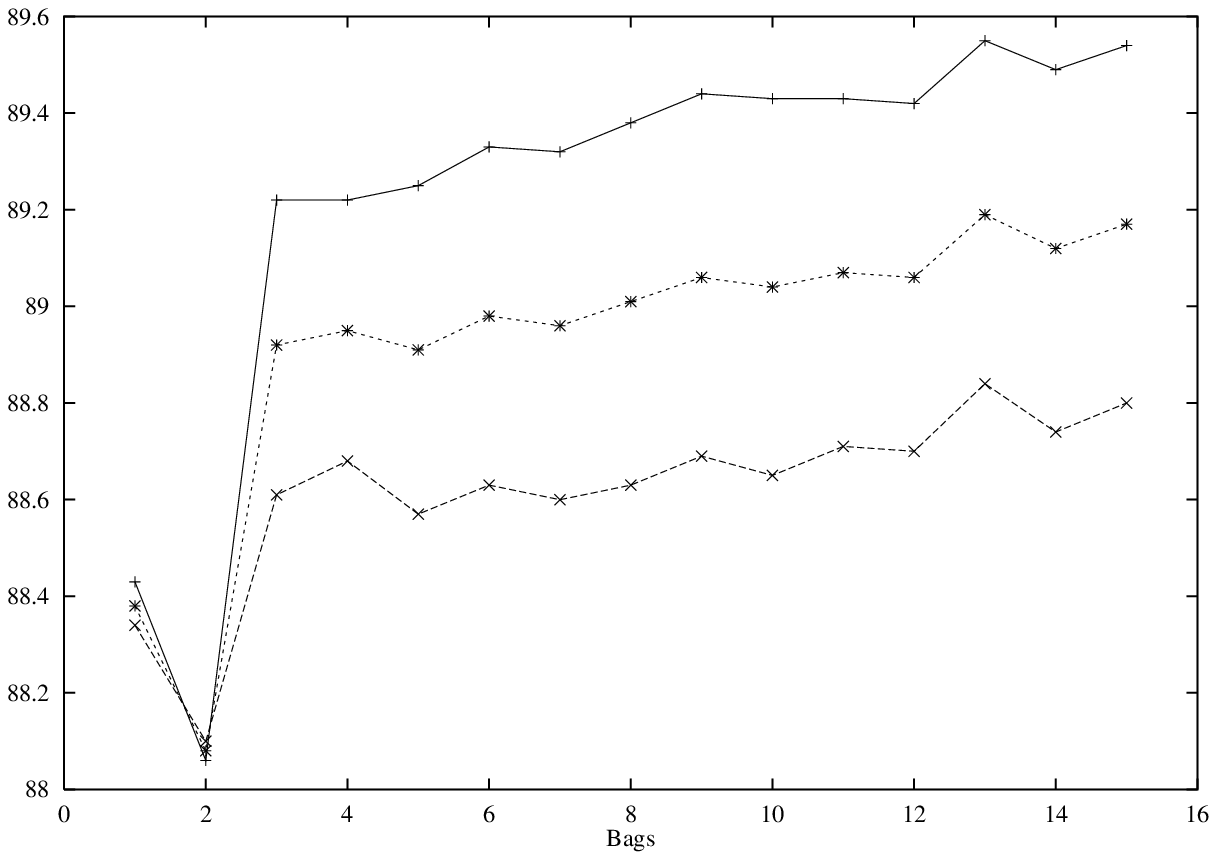, width= 0.25\textwidth}
&
    \epsfig{file=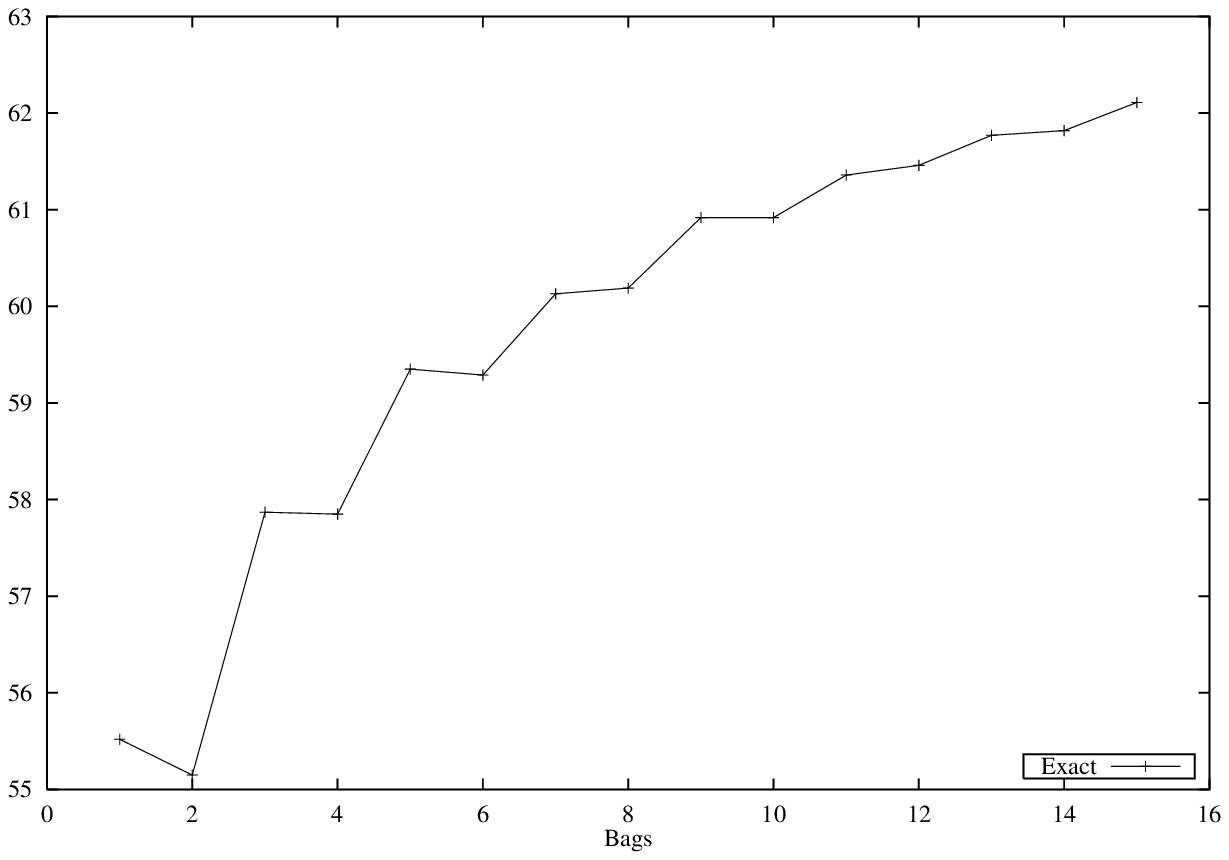, width= 0.25\textwidth}
&
    \epsfig{file=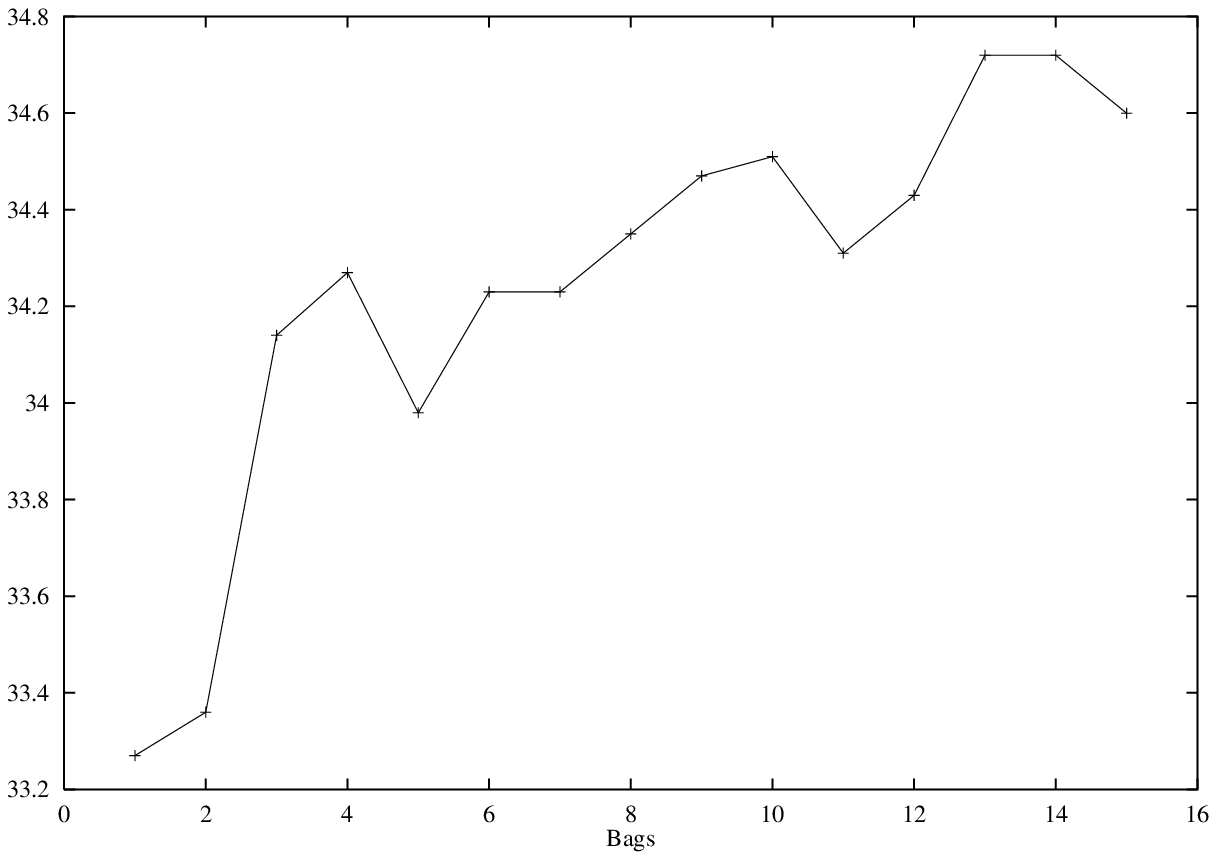, width= 0.25\textwidth}
\end{tabular}
}
%\\
%\bigskip

%%% Local Variables: 
%%% mode: latex
%%% TeX-master: "varying"
%%% End: 

%% file: boost.sent.fig.tex
%auto-ignore 

%\begin{figure}[htbp]
\mbox{
%\begin{tabular}{c}
\begin{tabular}{cccc}
    \epsfig{file=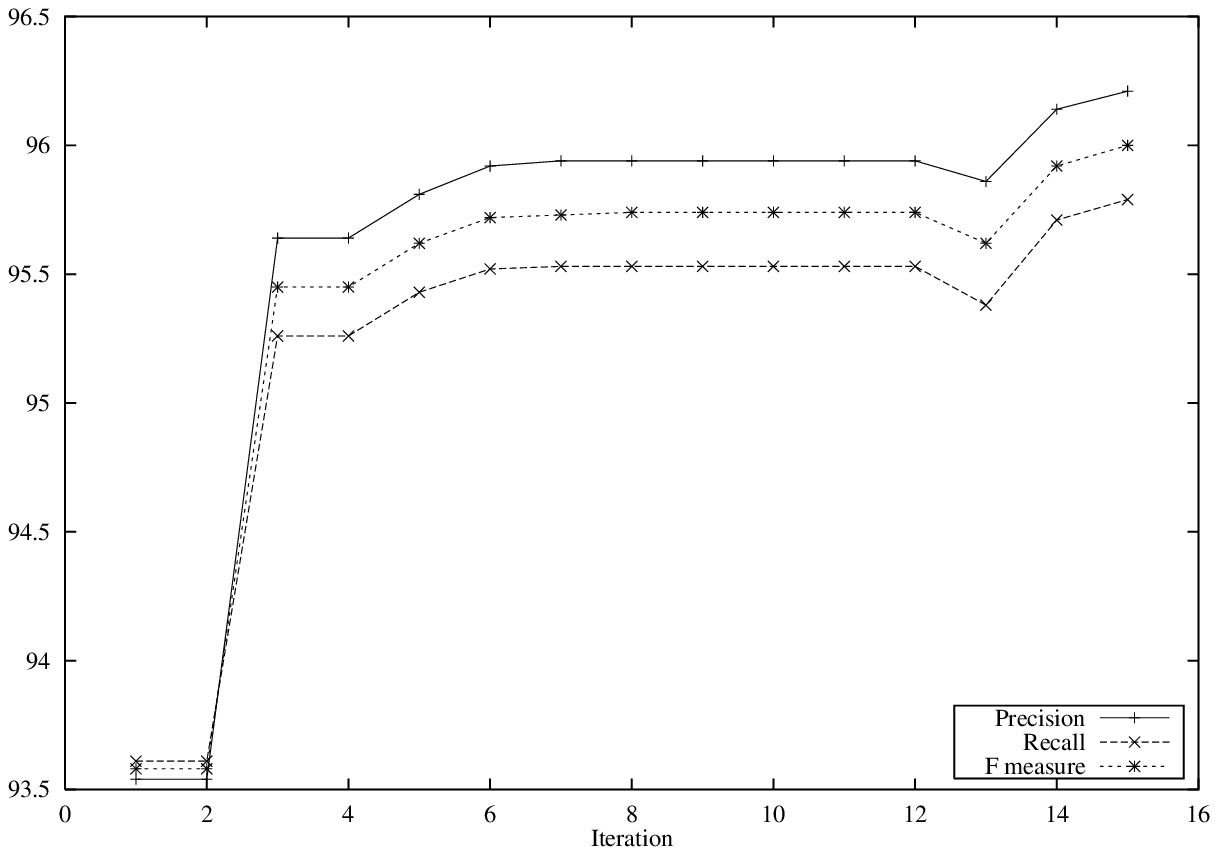, width= 0.25\textwidth}
&
    \epsfig{file=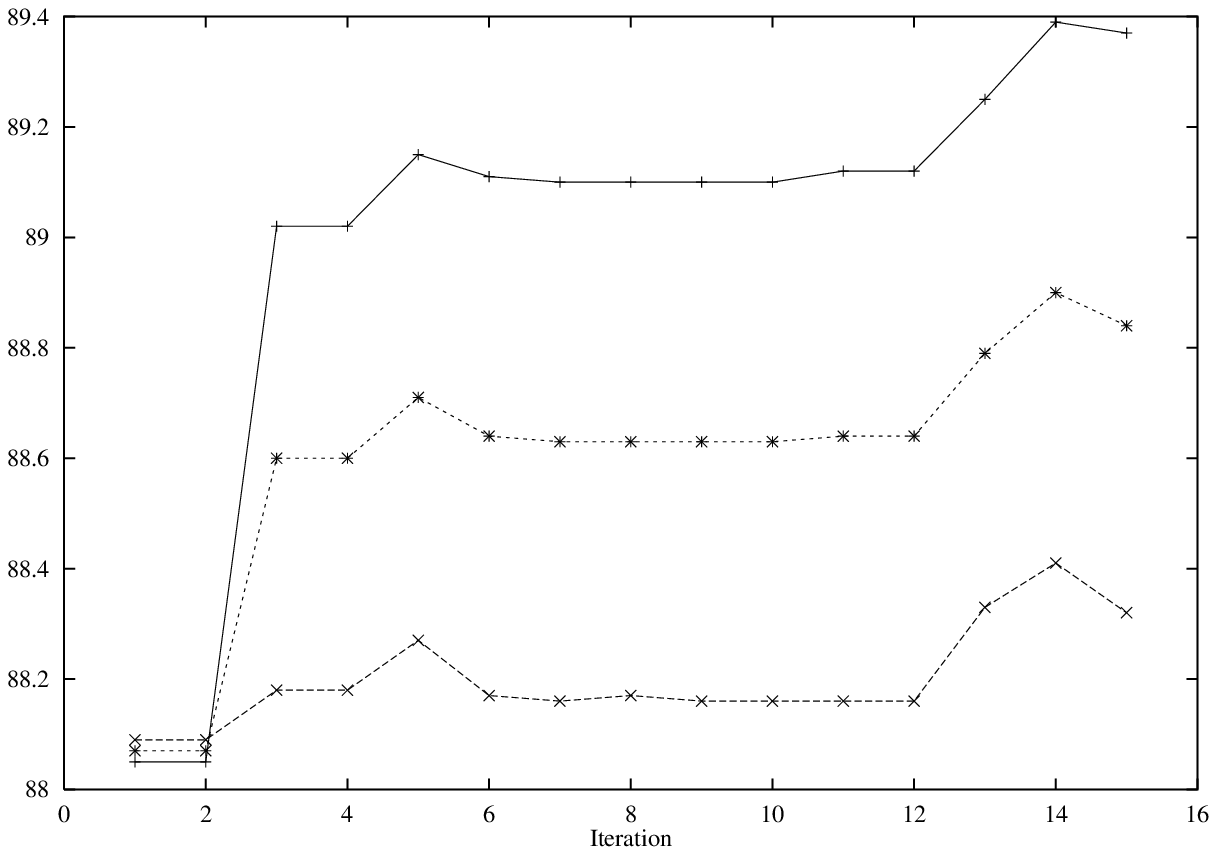, width= 0.25\textwidth}
&
    \epsfig{file=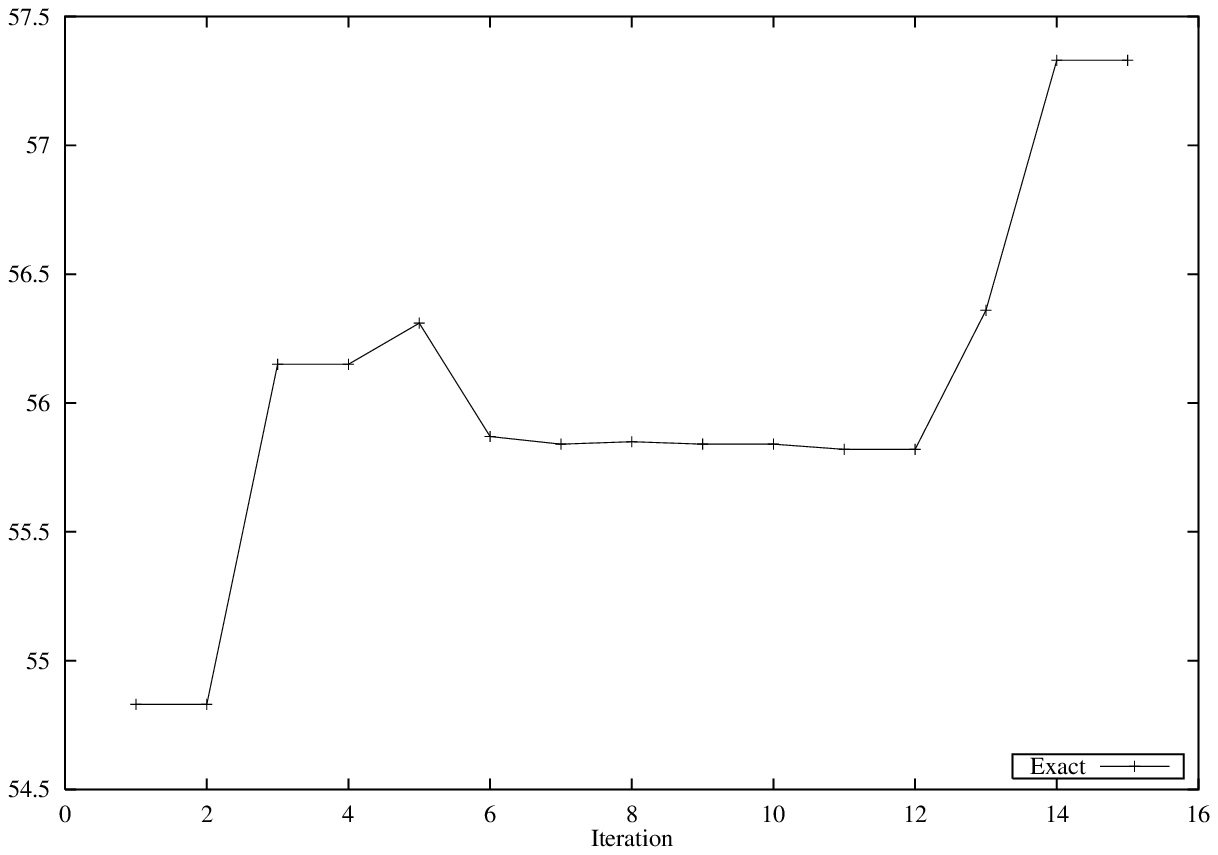, width= 0.25\textwidth}
&
    \epsfig{file=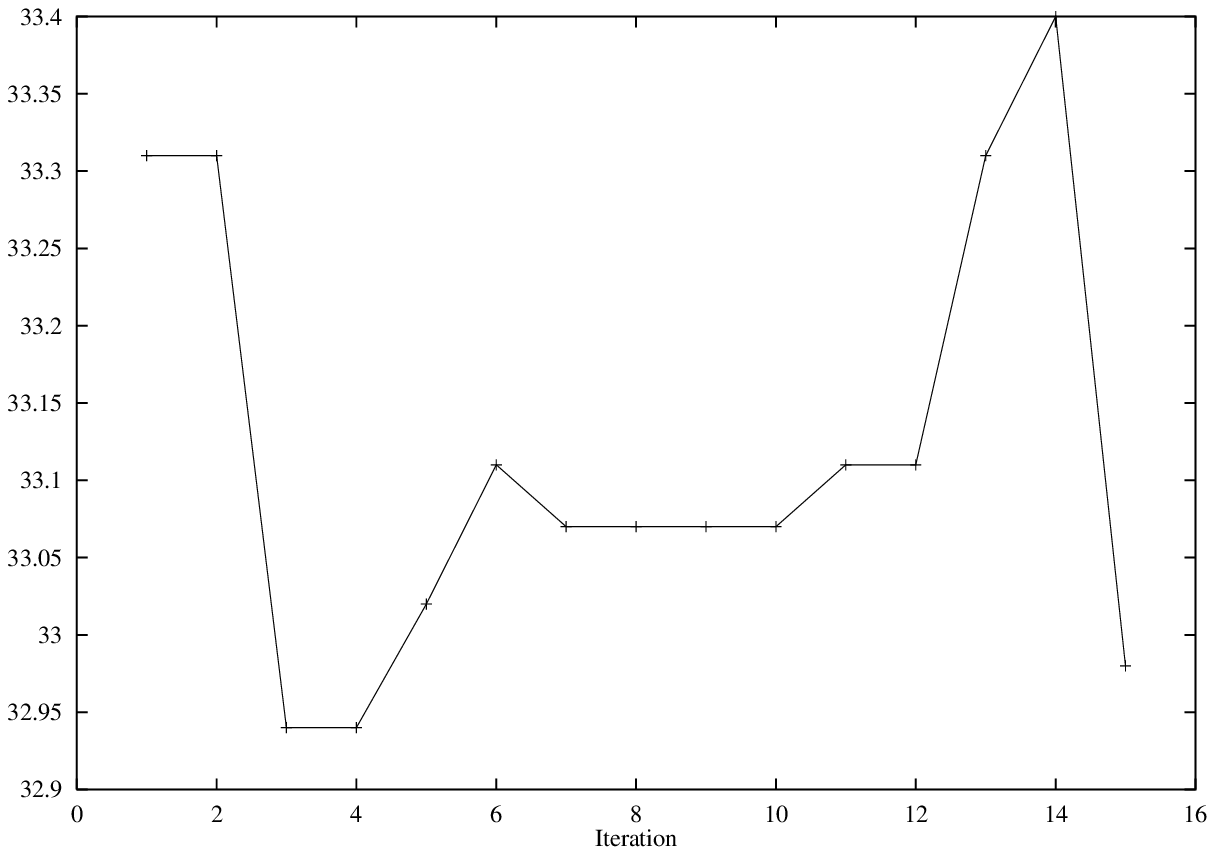, width= 0.25\textwidth}
\end{tabular}
}
%\\
%\bigskip

%%% Local Variables: 
%%% mode: latex
%%% TeX-master: "varying"
%%% End: 

%% file: sizes.vary.tex
%auto-ignore

%%\begin{figure}[htbp]
\mbox{
\begin{tabular}{cccc}
    \epsfig{file=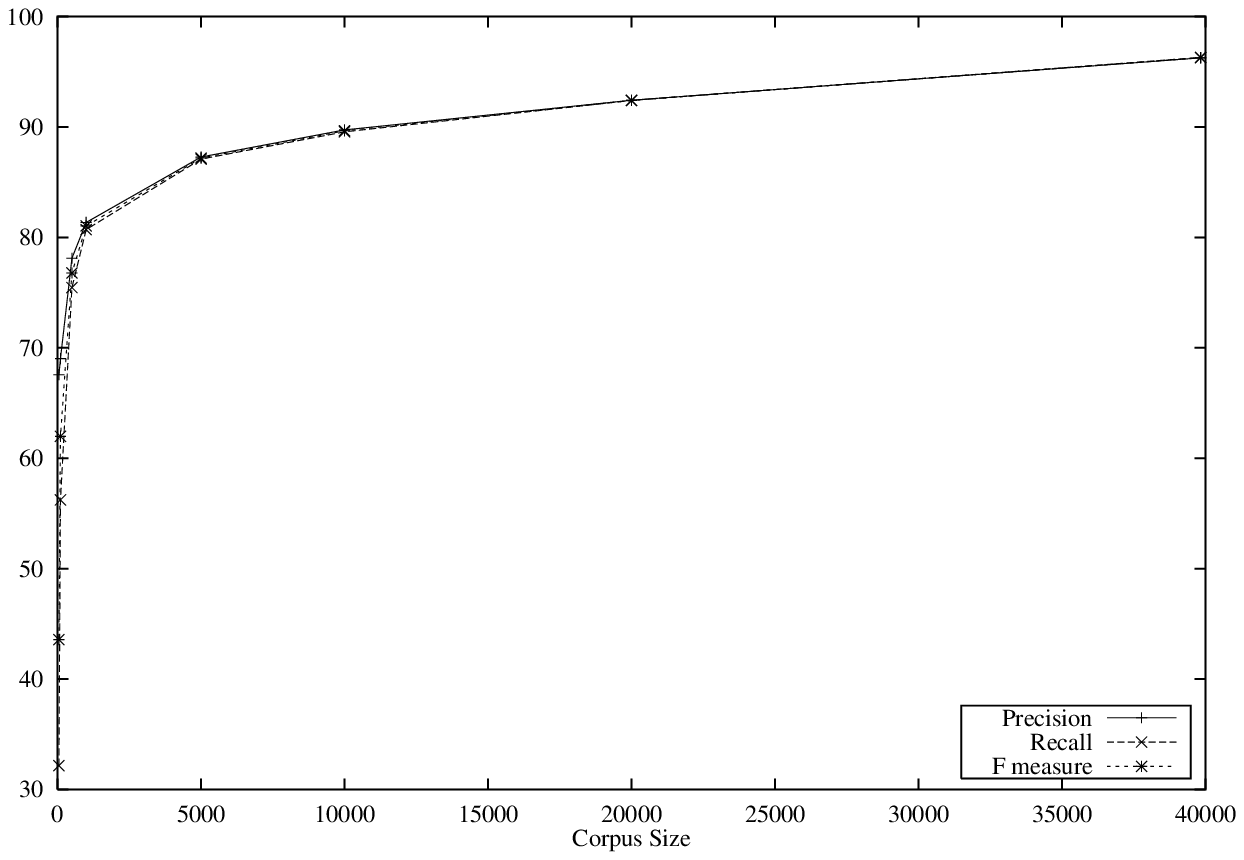, width= 0.25\textwidth}
&
    \epsfig{file=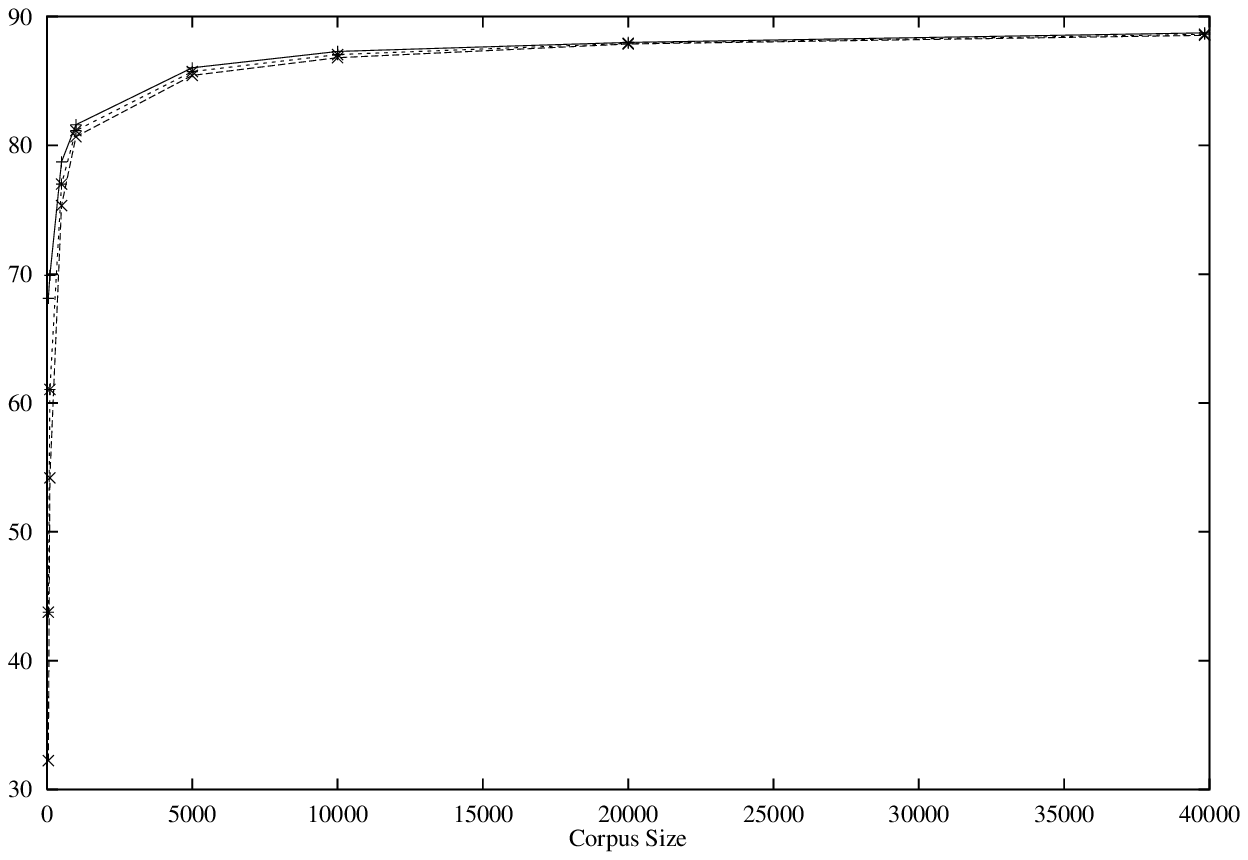, width= 0.25\textwidth}
&
    \epsfig{file=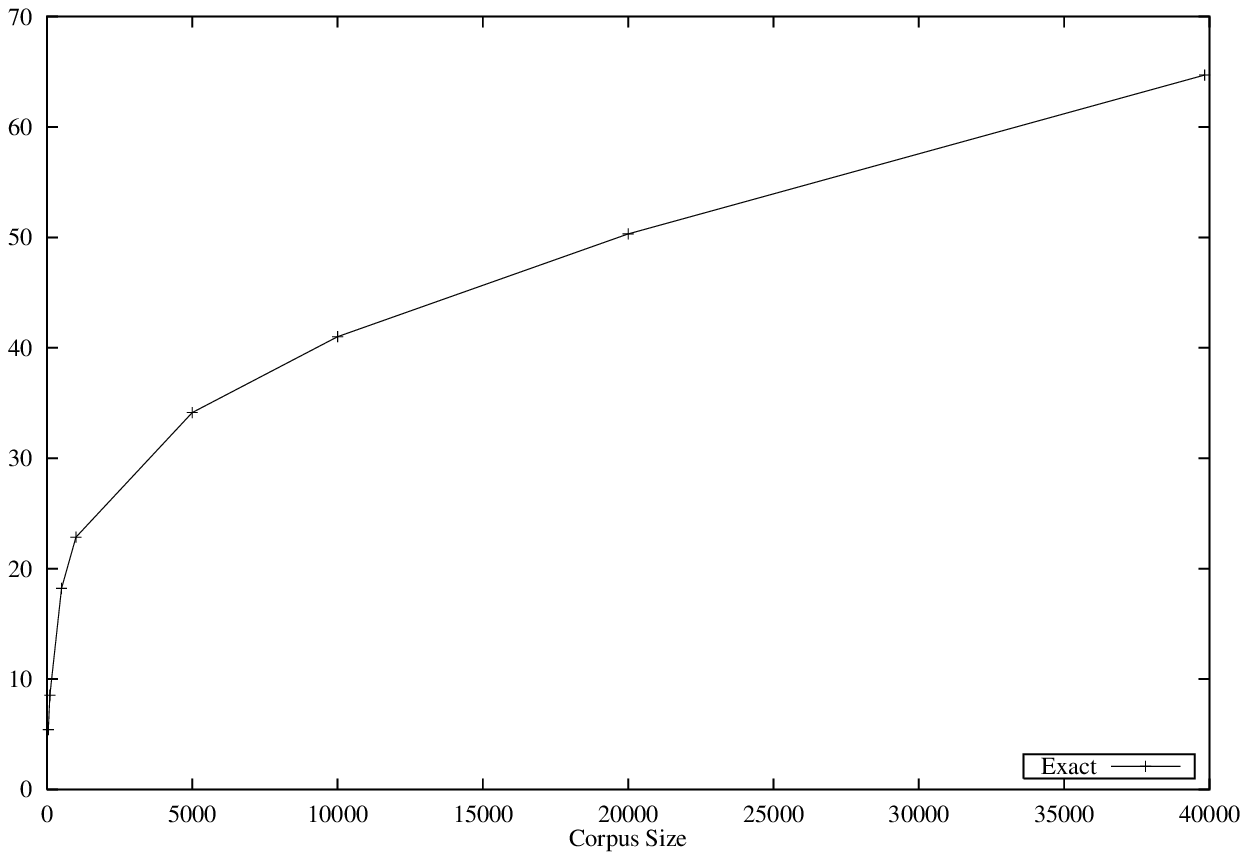, width= 0.25\textwidth}
&
    \epsfig{file=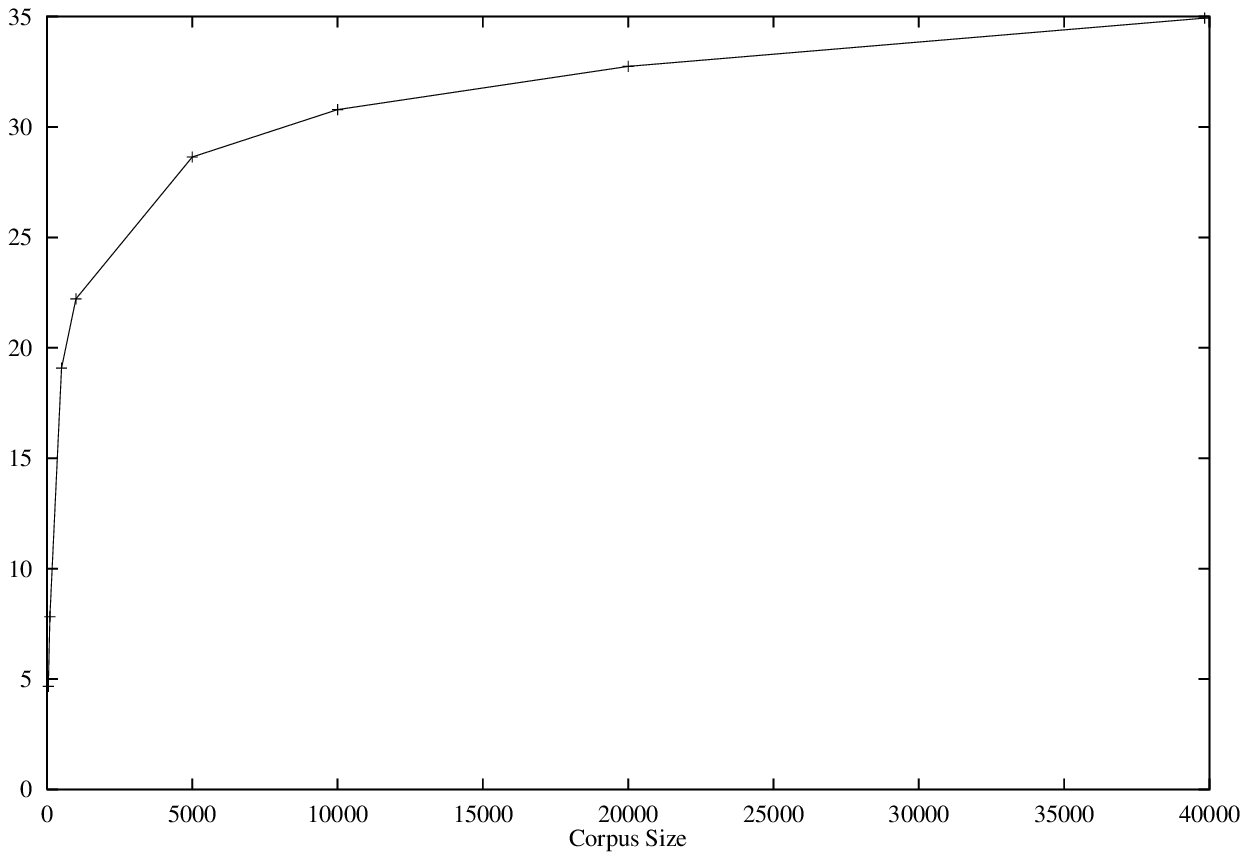, width= 0.25\textwidth}
\end{tabular}
}
%\end{figure}

%%% Local Variables: 
%%% mode: latex
%%% TeX-master: "varying"
%%% End: 

%% file: treebankerrors.tex
%auto-ignore

% \begin{parse}
% \label{firstparse}
% \hspace*{-\fill}
% \Tree
% [.TOP [.NP [.NNS Fees ]  [.CD 1 ]  [.CD 7/8 ]  ]  ] 
% \Tree
% [.TOP [.NP [.NP [.NNS Fees ]  ]  [.NP [.QP [.CD 1 ]  [.CD 7/8 ]  ]  ]  ]  ] 
% \Tree
% [.TOP [.NP [.NP [.NNS Fees ]  ]  [.NP [.CD 1 ]  [.CD 7/8 ]  ]  ]  ] 
% \Tree
% [.TOP [.NP [.NNS Fees ]  [.CD 1 ]  [.CD 3/8 ]  ]  ] 
% \Tree
% [.TOP [.NP [.NNS Fees ]  [.QP [.CD 1 ]  [.CD 3/8 ]  ]  ]  ] 
% \Tree
% [.TOP [.NP [.NP [.NNS Fees ]  ]  [.NP [.CD 1 ]  [.CD 3/8 ]  ]  ]  ] 
% \Tree
% [.TOP{\footnotemark}
%  [.NP [.NNS Fees ]  [.QP [.CD 1 ]  [.CD 7/8 ]  ]  ]  ] 
% \footnotetext{4 copies of this tree appeared in the set.}
% \begin{comment}
% \end{comment}
% \end{parse} 

% \newpage

% \begin{parse}
% \Tree
% [.TOP [.PP [.IN In ]  [.NP [.JJ other ]  [.NN commodity ]  [.NNS
% markets ]  [.NN yesterday ]  ]  ]  ]  
% \end{parse} 

% \begin{parse}
% \Tree
% [.TOP [.UCP [.PP [.IN In ]  [.NP [.JJ other ]  [.NN commodity ]  [.NNS
% markets ]  ]  ]  [.NP [.NN yesterday ]  ]  ]  ]  
% \end{parse} 

% \begin{parse}
% \Tree
% [.TOP{\footnotemark}
%  [.FRAG [.PP [.IN In ]  [.NP [.JJ other ]  [.NN commodity ]
% [.NNS markets ]  ]  ]  [.NP [.NN yesterday ]  ]  ]  ]  
% \footnotetext{5 copies.}
% \end{parse} 

%\onecolumn

\begin{sidewaysfigure}[htbp]
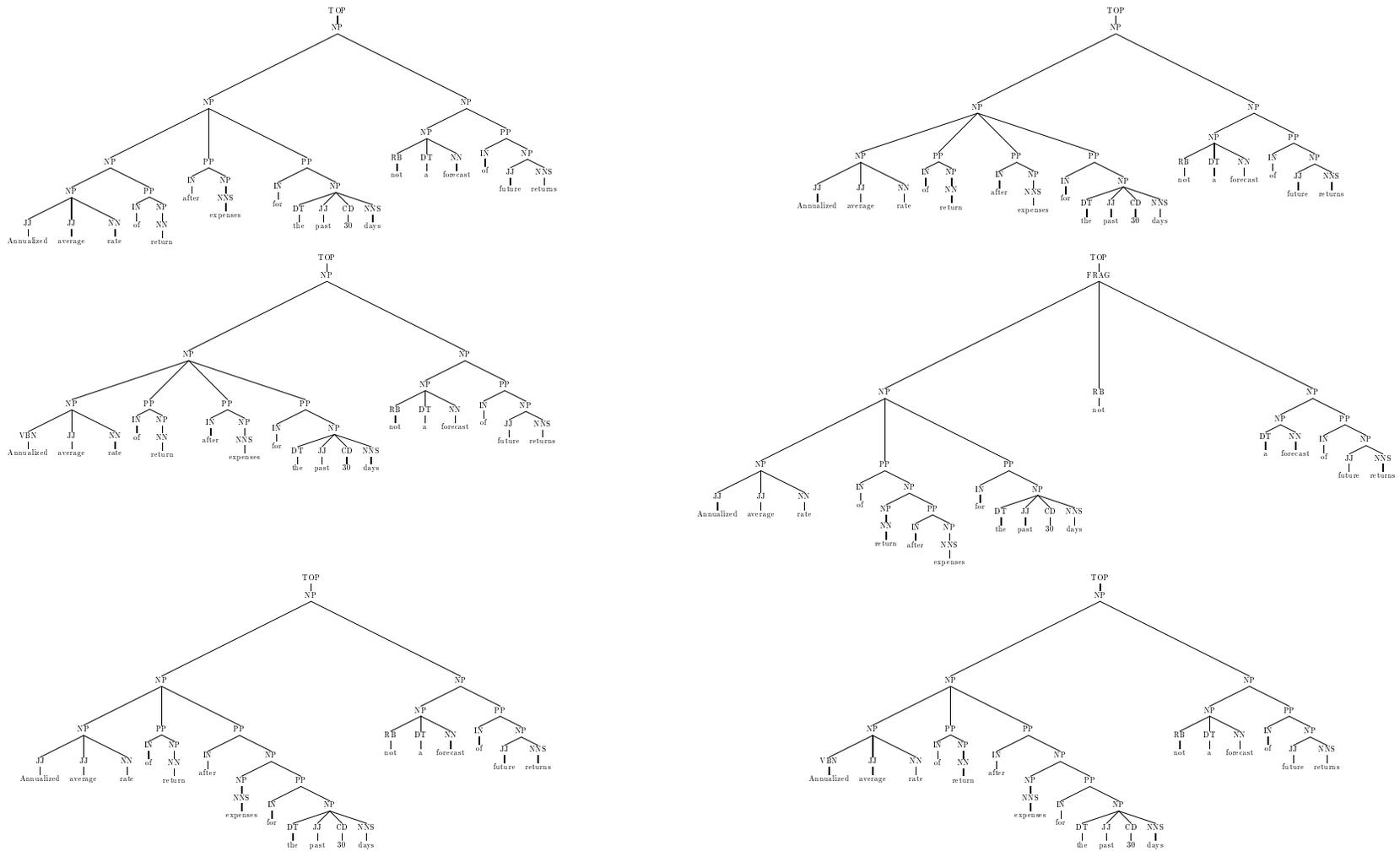

\scalebox{.37}{
\begin{tabular}{cc}
\Tree [.TOP [.NP [.NP [.NP [.NP [.JJ
Annualized ]  [.JJ average ]  [.NN rate
]  ]  [.PP [.IN of ]  [.NP [.NN return ]  ]  ]  ]  [.PP [.IN after ]
[.NP [.NNS expenses ]  ]  ]  [.PP [.IN for ]  [.NP [.DT the ]  [.JJ
past ]  [.CD 30 ]  [.NNS days ]  ]  ]  ]  [.NP [.NP [.RB not ]  [.DT a
]  [.NN forecast ]  ]  [.PP [.IN of ]  [.NP [.JJ future ]  [.NNS
returns ]  ]  ]  ]  ]  ] 
&
\Tree
[.TOP [.NP [.NP [.NP [.JJ Annualized ]  [.JJ average ]  [.NN rate ]  ]
[.PP [.IN of ]  [.NP [.NN return ]  ]  ]  [.PP [.IN after ]  [.NP
[.NNS expenses ]  ]  ]  [.PP [.IN for ]  [.NP [.DT the ]  [.JJ past ]
[.CD 30 ]  [.NNS days ]  ]  ]  ]  [.NP [.NP [.RB not ]  [.DT a ]  [.NN
forecast ]  ]  [.PP [.IN of ]  [.NP [.JJ future ]  [.NNS returns ]  ]
]  ]  ]  ]  
\\
\Tree
[.TOP [.NP [.NP [.NP [.VBN Annualized ]  [.JJ average ]  [.NN rate ]
]  [.PP [.IN of ]  [.NP [.NN return ]  ]  ]  [.PP [.IN after ]  [.NP
[.NNS expenses ]  ]  ]  [.PP [.IN for ]  [.NP [.DT the ]  [.JJ past ]
[.CD 30 ]  [.NNS days ]  ]  ]  ]  [.NP [.NP [.RB not ]  [.DT a ]  [.NN
forecast ]  ]  [.PP [.IN of ]  [.NP [.JJ future ]  [.NNS returns ]  ]
]  ]  ]  ]  
&
\Tree
[.TOP [.FRAG [.NP [.NP [.JJ Annualized ]  [.JJ average ]  [.NN rate ]
]  [.PP [.IN of ]  [.NP [.NP [.NN return ]  ]  [.PP [.IN after ]  [.NP
[.NNS expenses ]  ]  ]  ]  ]  [.PP [.IN for ]  [.NP [.DT the ]  [.JJ
past ]  [.CD 30 ]  [.NNS days ]  ]  ]  ]  [.RB not ]  [.NP [.NP [.DT a
]  [.NN forecast ]  ]  [.PP [.IN of ]  [.NP [.JJ future ]  [.NNS
returns ]  ]  ]  ]  ]  ]  
\\
\Tree
[.TOP [.NP [.NP [.NP [.JJ Annualized ]  [.JJ average ]  [.NN rate ]  ]
[.PP [.IN of ]  [.NP [.NN return ]  ]  ]  [.PP [.IN after ]  [.NP [.NP
[.NNS expenses ]  ]  [.PP [.IN for ]  [.NP [.DT the ]  [.JJ past ]
[.CD 30 ]  [.NNS days ]  ]  ]  ]  ]  ]  [.NP [.NP [.RB not ]  [.DT a ]
[.NN forecast ]  ]  [.PP [.IN of ]  [.NP [.JJ future ]  [.NNS returns
]  ]  ]  ]  ]  ]  
&
\Tree
[.TOP [.NP [.NP [.NP [.VBN Annualized ]  [.JJ average ]  [.NN rate ]
]  [.PP [.IN of ]  [.NP [.NN return ]  ]  ]  [.PP [.IN after ]  [.NP
[.NP [.NNS expenses ]  ]  [.PP [.IN for ]  [.NP [.DT the ]  [.JJ past
]  [.CD 30 ]  [.NNS days ]  ]  ]  ]  ]  ]  [.NP [.NP [.RB not ]  [.DT
a ]  [.NN forecast ]  ]  [.PP [.IN of ]  [.NP [.JJ future ]  [.NNS
returns ]  ]  ]  ]  ]  ]  
\end{tabular}
}
\caption{An inconsistent set of annotations from the Treebank}
\label{fig:treebankerrors}
\end{sidewaysfigure}

% \newpage

% \begin{sideways}
% \begin{parse}
% \scalebox{.7}{
% \Tree
% [.TOP [.NP [.NP [.JJ FEDERAL ]  [.NNS FUNDS ]  ]  [.NP [.NP [.ADJP [.CD 9 ]  [.NN (\%) ]  ]  [.JJ high ]  ]  [.NP [.ADJP [.QP [.CD 8 ]  [.CD 13/16 ]  ]  [.NN (\%) ]  ]  [.JJ low ]  ]  [.NP [.ADJP [.QP [.CD 8 ]  [.CD 7/8 ]  ]  [.NN (\%) ]  ]  [.ADJP [.IN near ]  [.NN closing ]  ]  [.NN bid ]  ]  [.NP [.NP [.QP [.CD 8 ]  [.CD 15/16 ]  ]  [.NN (\%) ]  ]  [.VP [.VBN offered ]  ]  ]  ]  ]  ] 
% }
% \end{parse}
% \end{sideways}

% \begin{sideways}
% \begin{parse}
% \scalebox{.7}{
% \Tree
% [.TOP [.NP [.NP [.JJ FEDERAL ]  [.NNS FUNDS ]  ]  [.NP [.NP [.NP [.QP [.CD 8 ]  [.CD 3/4 ]  ]  [.NN (\%) ]  ]  [.ADJP [.JJ high ]  ]  ]  [.NP [.NP [.QP [.CD 8 ]  [.CD 5/8 ]  ]  [.NN (\%) ]  ]  [.ADJP [.JJ low ]  ]  ]  [.NP [.NP [.QP [.CD 8 ]  [.CD 11/16 ]  ]  [.NN (\%) ]  ]  [.PP [.IN near ]  [.NP [.NN closing ]  ]  ]  [.VP [.NN bid ]  ]  ]  [.NP [.NP [.QP [.CD 8 ]  [.CD 11/16 ]  ]  [.NN (\%) ]  ]  [.VP [.VBN offered ]  ]  ]  ]  ]  ] 
% }
% \end{parse}
% \end{sideways}

% \begin{sideways}
% \begin{parse}
% \label{lastparse}
% \scalebox{.7}{
% \Tree
% [.TOP [.NP [.NP [.JJ FEDERAL ]  [.NNS FUNDS ]  ]  [.NP [.NP [.NP [.CD 8 ]  [.CD 3/4 ]  [.NN (\%) ]  ]  [.ADJP [.JJ high ]  ]  ]  [.NP [.NP [.CD 8 ]  [.CD 5/8 ]  [.NN (\%) ]  ]  [.ADJP [.JJ low ]  ]  ]  [.NP [.NP [.CD 8 ]  [.CD 11/16 ]  [.NN (\%) ]  ]  [.PP [.IN near ]  [.NP [.NN closing ]  [.NN bid ]  ]  ]  ]  [.NP [.NP [.CD 8 ]  [.CD 3/4 ]  [.NN (\%) ]  ]  [.VP [.VBN offered ]  ]  ]  ]  ]  ] 
% }
% \end{parse}
% \end{sideways}

%%% Local Variables: 
%%% mode: latex
%%% TeX-master: "varying"
%%% End: 